\documentclass[]{TEAI}
\usepackage{array}

\title{FlowTranslator: Training-Free In-Image Machine Translation with Flow-Based Editing}

\author[1,2]{Zhixian Li}
\author[1,2]{Xingsong Ye}
\author[1,2]{Yongkun Du}
\author[1,2]{Yunyan Xiao}
\author[1,2]{Yanglin Huang}
\author[1,2]{Pinxuan Chen}
\author[1,2]{Jiyu Zhuang}
\author[1,2]{Cong Xu}
\author[1,2,\dagger]{Zhineng Chen}

\affiliation[1]{\mbox{Institute of Trustworthy Embodied AI, Fudan University}}
\affiliation[2]{\mbox{Shanghai Key Laboratory of Multimodal Embodied AI}}

\contribution[\dagger]{Corresponding author}

\abstract{

In-image machine translation (IIMT) replaces text in an image with its translation while preserving the surrounding visual context. Existing task-specific models achieve convincing performance but require massive paired training images before and after translation. As advanced text-to-image (T2I) models have reached a practically usable level of multilingual text generation, we argue that this capability can be transferred to IIMT and present FlowTranslator. It is a training-free IIMT framework that first employs an OCR model and an LLM to generate the translation prompt, which is injected into the T2I-based diffusion model. To optimize the accurate and natural generation of translated text, we introduce two trajectory controls: Region Coupling Release (RCR) partially decorrelates target-branch noise inside the text region early in the trajectory, while Coupling Expansion (CE) amplifies the emerging edit displacement mid-trajectory. Under the EasyOCR read-back protocol, FlowTranslator achieves average PRIM BLEU and COMET scores of 20.6 and 57.2, respectively, substantially outperforming the strong training-based baseline UniTranslator. On the more challenging and realistic whole-image ScenePair photographs translated into Chinese, it reaches 40.1\% exact-match accuracy and surpasses existing editing systems, some of which only address the text-rendering stage rather than the complete IIMT task. These results establish flow-based editing as a strong, fully training-free baseline for IIMT. 

}
\correspondence{\email{zhinchen@fudan.edu.cn}}

\begin{document}
\maketitle
\section{Introduction}

\begin{figure}[t]
\centering
\includegraphics[width=0.85\textwidth]{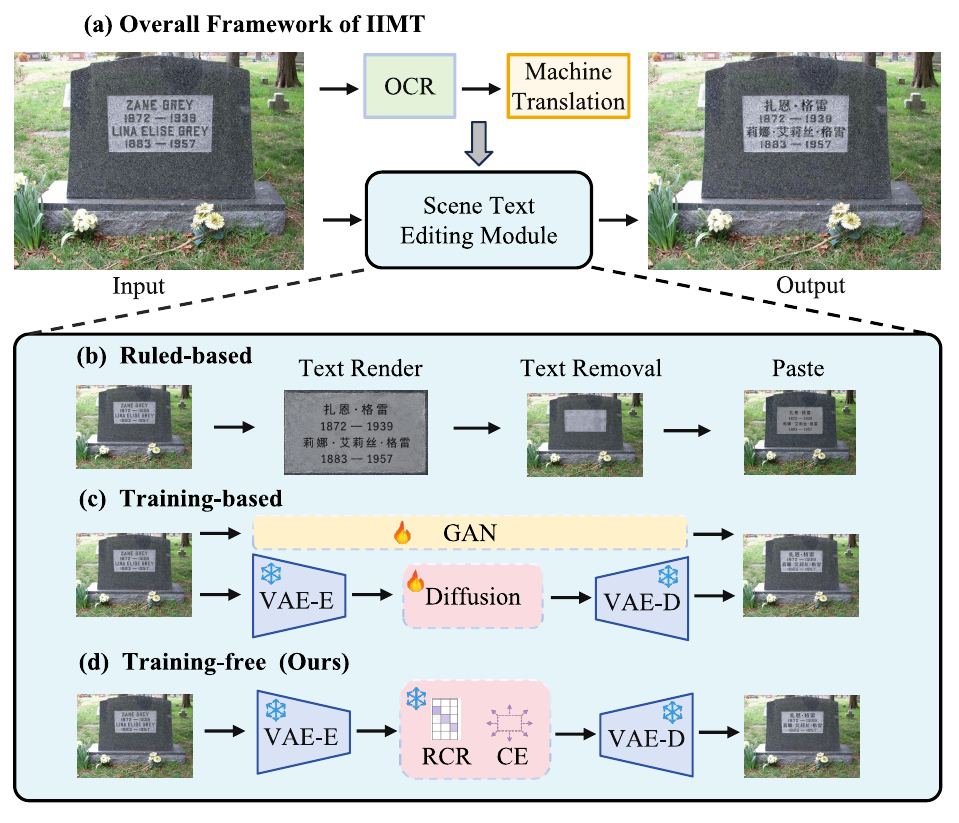}
\caption{Overview of the IIMT framework and editor paradigms. (a) The pipeline: OCR extracts text → a translator produces the translation → a scene text editor generates the output image. (b) Rule-based: remove, render, and paste separately. (c) Training-based: a learned GAN or diffusion editor (requires extra training). (d) FlowTranslator (ours): training-free flow editing with trajectory controls (CE and RCR) for text translation with background preservation.}
\label{fig:method-comparison}
\end{figure}

In-image machine translation (IIMT) translates the text embedded in an image and renders the translation back into the original text regions while keeping the rest of the image intact~\cite{mansimov2020towards,lan2024translatotron}.
The task underpins applications from mobile photo translation to localizing signage and product imagery, and has attracted dedicated benchmarks and models~\cite{tian2025exploring,tian2025prim,unitranslator2026}. A dominant line of work approaches IIMT with a cascaded pipeline: an OCR model localizes and reads the source text, a translation model converts it, and a rendering module writes the translation back into the image~\cite{zhu2023peit,lan2023exploring}. With the rapid advance of OCR and machine translation, the remaining bottleneck for this task is the rendering step.

Early rendering systems erase the source glyphs and paste separately rendered text, a procedure that leaves visible seams, mismatched typography, and destroyed textures. Subsequent work replaces the paste step with a learned scene-text-editing (STE) generator, first GAN-based~\cite{wu2019editingwild,mostel} and then diffusion-based~\cite{anytext,textdiffuser,ye2025textssr}, trained to blend the translated glyphs into the surrounding texture. However, these generators need paired images for training. Such cross-lingual pairs are scarce beyond a handful of high-resource directions, so training pipelines fall back on synthetically rendered text~\cite{tian2025prim}, which narrows the fonts, layouts, and backgrounds the generator ever sees and leaves the fragile rendering step as the bottleneck of the cascade.

Text-to-image (T2I) generation has meanwhile progressed to a point where multilingual text rendering is usable in practice: recent flow-matching backbones render legible non-Latin and CJK (Chinese-Japanese-Korean) text directly from a caption~\cite{zimage}. This raises a natural question. Rather than training another dedicated STE generator on scarce cross-lingual pairs, can a frozen T2I model be repurposed, with no task-specific training, to perform the rendering step of an IIMT cascade?

Inversion-free flow editing offers exactly this machinery. FlowEdit~\cite{kulikov2025flowedit} edits a real image with a frozen T2I flow model by integrating an ODE between a source-conditioned branch and a target-conditioned branch, and TextFlow~\cite{textflow2026} shows that such training-free editing can match trained scene-text editors in visual quality. Both operate on a single word or crop given a user-provided target string. Neither addresses the full IIMT problem of localizing, translating, and re-rendering every text region in an image automatically.

Therefore, we propose \textbf{FlowTranslator}, a novel IIMT system that requires no task-specific training. Every module is a frozen, publicly documented pretrained model, and no parameter is updated or fine-tuned for IIMT (the components themselves are, of course, products of large-scale pretraining).
FlowTranslator keeps the modularity of a cascade. A general OCR model~\cite{ppocr} finds text blocks, and a frozen LLM translates them. The pipeline replaces the training-based rendering stage with region-masked flow editing on a frozen T2I model~\cite{zimage}. The translation is generated into the original image rather than pasted onto it, so the diffusion prior handles perspective, lighting, and typography.

Making this transfer work requires solving a problem specific to cross-lingual, mask-confined editing: an \textbf{ignition failure} of the underlying flow editor.
The FlowEdit update is driven by the velocity difference between the source-conditioned and target-conditioned branches, which share the same noise. The difference is meant to grow as the caption change propagates a semantic edit. FlowEdit is demonstrated on edits, such as swapping an object or changing a global style, that alter a substantial fraction of the image and so produce an early, sizeable velocity difference. IIMT edits invert this premise. The text region is typically a small minority of the frame, so the two branches barely differ at the outset and the edit never starts, no matter how the difference is subsequently amplified. This text-minority regime is intrinsic to cross-lingual scene-text editing rather than an artifact of our implementation, so a generic remedy such as increasing the velocity gain cannot repair it: there is no signal left to amplify.
We introduce two trajectory controls confined to the text-region mask to force ignition:
\textbf{Region Coupling Release} (RCR) replaces the target-branch noise inside the mask with a partially decorrelated sample early in the trajectory, giving the target branch room to diverge.
\textbf{Coupling Expansion} (CE) amplifies the nascent edit displacement mid-phase, letting the target branch complete the edit it has started.
Both add no forward passes and contribute zero latent update outside the mask.

We evaluate FlowTranslator on three benchmarks spanning synthetic and real imagery, seven languages, and nine translation directions. It achieves the highest average BLEU on Translatotron-V(ision)~\cite{lan2024translatotron} and the highest average BLEU and COMET on PRIM~\cite{tian2025prim} without any IIMT training data. On ScenePair photographs translated into Chinese~\cite{mostel}, FlowTranslator reaches 40.1\% exact-match accuracy and outperforms six released editing systems under a common evaluator. These results establish FlowTranslator as a strong training-free baseline across diverse IIMT settings.

In summary, our main contributions are:
\begin{itemize}
\item We repurpose a frozen, training-free text-to-image flow model to replace the training-based rendering stage of IIMT cascades and remove its dependence on scarce cross-lingual training pairs.
\item We identify an edit-ignition failure of coupled flow editing and propose two mask-confined trajectory controls (RCR and CE) without additional forward passes to resolve it.
\item We report results on TTV, PRIM, and ScenePair competitive with trained systems, establishing a strong training-free baseline for IIMT.
\end{itemize}
\section{Related Work}

\paragraph{In-image Machine Translation.}
Cascaded systems remain the dominant paradigm for IIMT, decomposing the task into OCR~\cite{ppocr,du2022svtr,du2025svtrv2,zheng2024cdistnet,ye2026all}, machine translation, and text rendering~\cite{zhu2023peit,lan2023exploring,xie2026decompose}. This modular design has advanced alongside its components: stronger OCR~\cite{du2026unirec,ye2026advancing,ye2026what} and translation models improve text recovery and translation, while the rendering stage has evolved from removing the source text and pasting separately rendered glyphs to learned scene-text editors that better preserve background, layout, and typography~\cite{mostel,anytext,textdiffuser}. Cascades can therefore incorporate strong pretrained models and upgrade each stage independently, but they still suffer from error propagation, and rendering often remains the principal source of visual degradation. End-to-end IIMT has been explored as an alternative, from early pixel-level models~\cite{mansimov2020towards} to Translatotron-V~\cite{lan2024translatotron}, VisTrans~\cite{tian2025prim}, and recent unified multimodal systems such as UniTranslator~\cite{unitranslator2026} and AnyTrans~\cite{qian2024anytrans}. These recent models have narrowed the gap and report performance competitive with cascaded baselines on established benchmarks, showing the growing viability of unified modeling. However, all published full-pipeline IIMT systems we are aware of train or fine-tune at least one component on IIMT data, whereas FlowTranslator updates no parameters at all.

\paragraph{Scene Text Editing.}
The rendering step of IIMT needs editors to replace words in images with GAN or diffusion models. Early systems transfer or regenerate text appearance with adaptive or stroke-level models~\cite{wu2019editingwild,roy2020stefann,yang2020swaptext,mostel}. Recent diffusion editors~\cite{anytext,textdiffuser,zhao2024udifftext,zeng2024textctrl,lan2025fluxtext,textriro}. Recognition-aware editors further use text recognition to improve content fidelity, as in RS-STE~\cite{fang2025rsste}. For cross-lingual rendering, however, these editors need paired images showing the same scene rendered in both scripts. Such pairs are scarce beyond a few high-resource directions, so training pipelines fall back on synthetically composited text~\cite{tian2025prim}, narrowing the appearance diversity seen in training. 
TextFlow~\cite{textflow2026} shows that training-free flow editing can match trained editors on word crops. Our work lifts training-free editing to the full IIMT task with automatic localization and translation.

\paragraph{Training-free Flow Editing.}
Flow matching provides a simulation-free training objective for continuous generative flows~\cite{lipman2022flowmatching}. FlowEdit~\cite{kulikov2025flowedit} constructs an inversion-free ODE between source and target text conditions of a pretrained flow-matching model~\cite{esser2024sd3,flux2024} and is the foundation of our renderer.
Other recent studies regularize rectified-flow trajectories or design training-free conditional samplers for image editing~\cite{flowalign,wang2024tamingrf,zheng2025lanpaint,zhu2025kvedit}, reflecting a shift from constructing new ODEs to controlling them. Our RCR and CE belong to this family as spatially masked controls designed specifically for the text-minority ignition setting.

\section{Method}
\label{sec:method}

\begin{figure*}[t]
\centering
\includegraphics[width=0.95\textwidth]{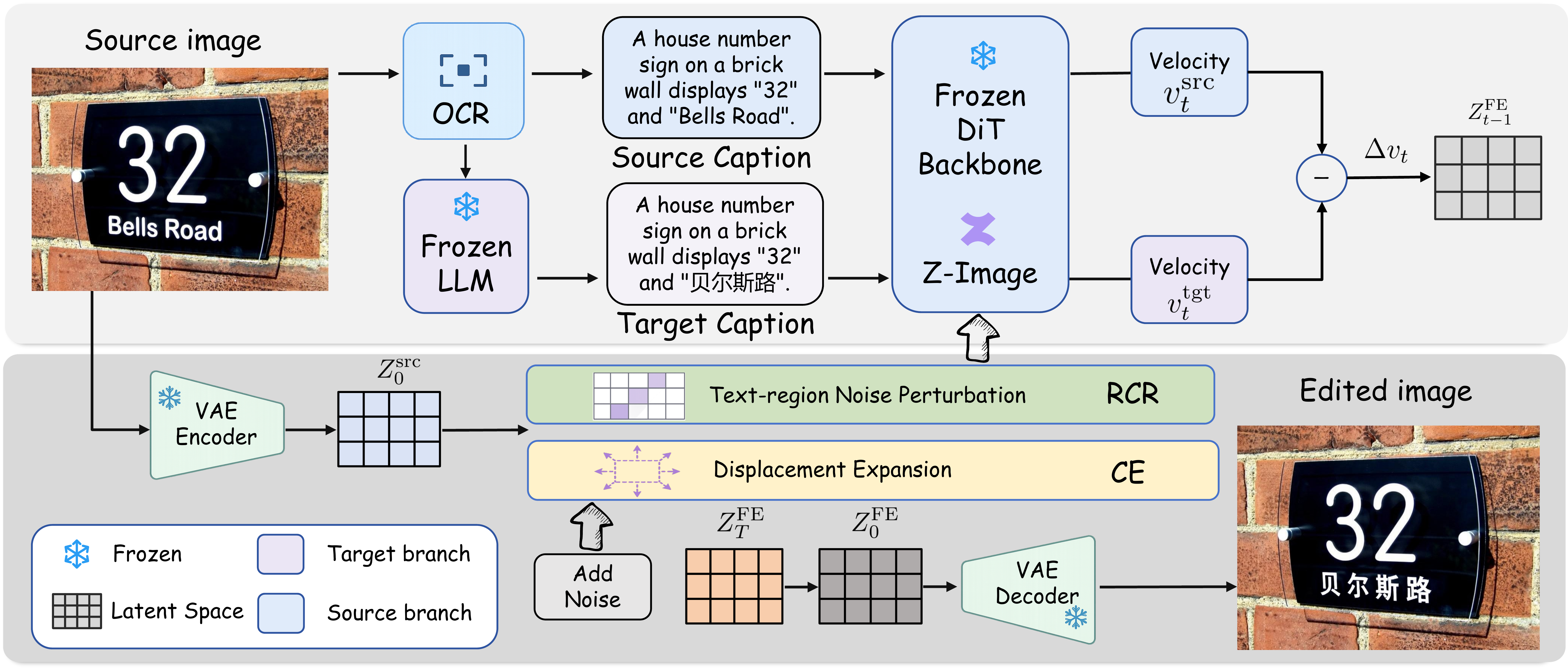}
\caption{Overview of FlowTranslator. OCR extracts the source text and forms a source caption, while a frozen LLM translates it to obtain the target caption. A frozen VAE encodes the source image into $Z^{\mathrm{src}}_0$, from which noise initialization produces $Z^{\mathrm{FE}}_T$. At each editing step, the same frozen DiT predicts source- and target-conditioned velocities. Their difference updates the flow-editing latent. RCR perturbs the target-branch noise within the detected text region, and CE amplifies the accumulated displacement before it is fed back to the target branch. The final latent $Z^{\mathrm{FE}}_0$ is decoded to produce the translated image.}
\label{fig:pipeline}
\end{figure*}

This section describes how FlowTranslator turns a frozen T2I model into a renderer for in-image machine translation, following the pipeline shown in Fig.~\ref{fig:pipeline}. 
FlowTranslator takes FlowEdit~\cite{kulikov2025flowedit}, an inversion-free editing procedure for pretrained flow-matching models, as its starting point.  It then adds two trajectory controls that overcome an ignition failure particular to cross-lingual, mask-confined editing, so that multilingual text is rendered accurately without task-specific training or extra inference cost.
No component in the pipeline is trained or fine-tuned.

\subsection{Overall Framework}
\label{sec:framework}

\paragraph{Input Preparation.}
Given a source image $x_{\mathrm{src}}$, the OCR model returns text boxes $\{B_i\}_{i=1}^{K}$ and recognized strings. The LLM translates each string and places the source and target strings in two otherwise matched scene descriptions, yielding $c_{\mathrm{src}}$ and $c_{\mathrm{tgt}}$. The image is encoded once by the frozen VAE ($Z^{\mathrm{src}}_0 = \mathcal{E}_{\mathrm{VAE}}(x_{\mathrm{src}})$), and the detected boxes are rasterized at latent resolution after dilation to form the editable mask:
\begin{equation}
\begin{aligned}
M &= \operatorname{Rasterize}_{\mathrm{lat}}\!\left(
\operatorname{Dilate}_{\delta}\!\left(\bigcup_{i=1}^{K} B_i\right)\right).
\end{aligned}
\label{eq:initialization}
\end{equation}
Here, $M$ has value one in the text region and zero elsewhere. The same mask is used for the ordinary FlowEdit update and for both trajectory controls introduced below, so every intervention stays spatially confined to the detected text.

\paragraph{Region-masked Flow Editing.}
Let $Z^{\mathrm{src}}_0$ be with velocity field $v_\theta(z, t, c)$, and let $c_{\mathrm{src}}, c_{\mathrm{tgt}}$ be the source and target captions produced by the LLM.
Starting from $Z^{\mathrm{FE}}_T = Z^{\mathrm{src}}_0$, our region-masked implementation integrates
\begin{equation}
\begin{aligned}
\Delta v_t &= v_\theta(\hat z^{\mathrm{tar}}_t, t, c_{\mathrm{tgt}})
 - v_\theta(\hat z^{\mathrm{src}}_t, t, c_{\mathrm{src}}), \\
Z^{\mathrm{FE}}_{t-\Delta t} &= Z^{\mathrm{FE}}_{t}
 - \Delta t\, \gamma M \odot \Delta v_t.
\end{aligned}
\label{eq:flowedit}
\end{equation}
where $\hat z^{\mathrm{src}}_t = (1-t) Z^{\mathrm{src}}_0 + t\,\epsilon$ with $\epsilon \sim \mathcal N(0, I)$, and the target branch is centered at the current edited latent while reusing the same noise realization:
$\hat z^{\mathrm{tar}}_t = Z^{\mathrm{FE}}_t + \hat z^{\mathrm{src}}_t - Z^{\mathrm{src}}_0$.
The time variable runs from the noisy end toward the data end, so $t$ decreases from one to zero over the $T$ solver steps. The source branch is a reference path from the source latent toward Gaussian noise. The target branch carries the same stochastic residual but is centered at $Z^{\mathrm{FE}}_t$. The shared noise couples the two branches, while their caption-dependent difference supplies the semantic edit direction. In this view, $Z^{\mathrm{FE}}_t$ is the only state that accumulates the edit, while the two DiT evaluations provide the local direction used to update it.
The gain $\gamma \ge 1$ scales the velocity difference inside $M$, and editing is restricted to the first $n_{\max}$ of $T$ steps.
Latents outside $M$ are never modified, so the background stays frozen in latent space.

\paragraph{Ignition Failure.}
The previous method relies on the source and target branches diverging enough for $\Delta v_t$ to carry a semantic edit signal. In IIMT, the two branches share noise and differ only in a small text region, so their inputs and conditional velocities stay nearly identical near $t \approx 1$. Consequently, $\Delta v_t$ and the displacement $Z^{\mathrm{FE}}_t-Z^{\mathrm{src}}_0$ stay near zero, and the source text remains unchanged. Increasing $\gamma$ cannot amplify a signal that is not there: zero displacement is a fixed point of the update, and escaping it requires modifying the coupling itself rather than the gain applied to it.

We address this fixed point with two trajectory controls confined to the mask $M$, each targeting a distinct requirement for ignition. Region Coupling Release first loosens the noise coupling between the branches early in the trajectory, giving the target branch room to diverge from the source. Coupling Expansion then amplifies the resulting displacement in the middle of the trajectory, giving the target branch a direction to follow once it can move.

\subsection{Region Coupling Release}
\label{sec:rcr}

Region Coupling Release (RCR) targets the root cause of the fixed point, the perfect noise correlation between the branches.
Instead of feeding the target branch the source noise $\epsilon$, RCR feeds it a partially decorrelated sample inside the mask.
With a schedule $s(t)$ ramping linearly from $1$ (for $t \ge \tau_{\mathrm{hi}}$) to $0$ (for $t \le \tau_{\mathrm{lo}}$), we draw an independent $\epsilon' \sim \mathcal N(0, I)$, fresh at every step and every noise sample like $\epsilon$, and form
\begin{equation}
\epsilon^{\mathrm{tar}} = \rho\, \epsilon + \sqrt{1-\rho^2}\, \epsilon', \qquad \rho = 1 - \kappa\, s(t),
\label{eq:rcr-noise}
\end{equation}
\begin{equation}
\hat z^{\mathrm{tar}}_t \leftarrow \hat z^{\mathrm{tar}}_t + M \odot t\, (\epsilon^{\mathrm{tar}} - \epsilon).
\label{eq:rcr-update}
\end{equation}
Inside the text region the target branch now sees noise that is only $\rho$-correlated with the source branch, so the two velocities are evaluated at genuinely different points and their difference stops vanishing. The edit has room to ignite.

Three design choices govern this mechanism.
First, the decorrelation is partial, $\kappa < 1$: Eq.~\eqref{eq:rcr-noise} preserves the marginal distribution of $\epsilon^{\mathrm{tar}}$ while interpolating its coupling strength, whereas full decorrelation would sever the branches and degrade the edit into mask-region regeneration.
Second, the perturbation in Eq.~\eqref{eq:rcr-update} is additive on top of the accumulated displacement rather than an attenuation of it, so it preserves existing edit progress instead of rescaling the update and shrinking the edit along with the noise, as ordinary guidance schemes do.
Third, the release follows a schedule: it is widest early, for $t \ge \tau_{\mathrm{hi}}$, where the trajectory is noise-dominated and divergence is cheap, and it closes completely for $t \le \tau_{\mathrm{lo}}$, restoring exact coupling so that late steps refine glyphs under the same conditions as vanilla FlowEdit.
Together these choices let RCR open a controlled window for divergence without discarding the coupling that grounds the edit in the source image.

\subsection{Coupling Expansion}
\label{sec:ce}

RCR opens room to ignite but supplies no direction of its own, since a decorrelated branch may drift toward any content consistent with the target caption.
The needed direction is already present in the trajectory: the accumulated displacement $Z^{\mathrm{FE}}_t - Z^{\mathrm{src}}_0$ points from the source toward the emerging edit.
Coupling Expansion (CE) feeds this displacement back into the target branch as positive feedback:
\begin{equation}
\hat z^{\mathrm{tar}}_t \leftarrow \hat z^{\mathrm{tar}}_t + c \sin^2(\pi t)\, M \odot (Z^{\mathrm{FE}}_t - Z^{\mathrm{src}}_0),
\label{eq:ce}
\end{equation}
so the target branch is evaluated at a point where the nascent edit already appears more advanced than it is. The conditional velocity then responds by completing that edit, and the loop closes on itself.

The $\sin^2$ gate confines this feedback to the middle of the trajectory for a structural reason. Early on, there is no displacement to expand because $Z^{\mathrm{FE}}_t \approx Z^{\mathrm{src}}_0$, so CE stays inert on its own and cannot inject a spurious direction before RCR has created one. Late in the trajectory, the gate closes again to keep runaway amplification from distorting glyph shapes during refinement.
The feedback is also self-limiting within the edit window, since $c \sin^2(\pi t) \le c$ and editing stops at $n_{\max}$, so the expansion never compounds without bound.

RCR and CE are complementary rather than redundant: RCR opens the door and CE walks the edit through it. Our subsequent ablation (\S\ref{sec:ablations}) shows that CE alone carries most of the gain, while RCR adds a further increment on top and composes positively with CE across most translation directions.

\section{Experiments}

\begin{table*}[t]
\centering
\caption{TTV benchmark with EasyOCR read-back and lowercased sacreBLEU. ``IIMT-tuned'' marks systems trained or fine-tuned on IIMT data. GPT-Image-1 and Qwen-Image are zero-shot prompted unified models. ``Masked FlowEdit'' is our baseline with identical hyperparameters and the trajectory controls disabled. Average columns are unweighted means over the four directions. For each metric, best in \textbf{bold}, second \underline{underlined}.}
\label{tab:ttv}
{\small
\setlength{\tabcolsep}{5pt}
\begin{tabular}{l c cccc cc}
\toprule
& & \multicolumn{4}{c}{BLEU$\uparrow$} & \multicolumn{2}{c}{Avg.} \\
\cmidrule(lr){3-6}\cmidrule(lr){7-8}
Model & IIMT-tuned & De$\rightarrow$En & En$\rightarrow$De & Fr$\rightarrow$En & Ro$\rightarrow$En & BLEU$\uparrow$ & SSIM$\uparrow$ \\
\midrule
GPT-Image-1~\cite{openai2025gptimage} &  & 20.75 & \textbf{19.61} & \underline{27.38} & \textbf{24.50} & \underline{23.06} & 0.734 \\
Qwen-Image~\cite{wu2025qwenimage} &  & 0.23 & 0.21 & 0.20 & 0.22 & 0.22 & \underline{0.785} \\
\midrule
OCR+MT+T2I~\cite{lan2024translatotron} & \checkmark & 15.37 & 13.22 & 21.60 & 18.34 & 17.13 & 0.771 \\
PEIT+T2I~\cite{zhu2023peit} & \checkmark & 10.91 & 8.77 & 18.51 & 14.54 & 13.18 & 0.769 \\
Translatotron-V~\cite{lan2024translatotron} & \checkmark & 15.39 & 13.23 & 22.20 & 18.44 & 17.32 & 0.776 \\
UniTranslator~\cite{unitranslator2026} & \checkmark & \underline{25.03} & 13.41 & \textbf{27.77} & 18.45 & 21.17 & \textbf{0.805} \\
\midrule
Masked FlowEdit~\cite{kulikov2025flowedit} &  & 24.43 & \underline{17.53} & 26.89 & 22.84 & 22.92 & 0.754 \\
FlowTranslator (Ours) &  & \textbf{25.15} & 17.15 & 27.07 & \underline{23.37} & \textbf{23.19} & 0.746 \\
\bottomrule
\end{tabular}}

\end{table*}

\subsection{Evaluation and Implementation Details}
\label{sec:setup}

\paragraph{Benchmarks.}
\textbf{TTV} (Translatotron-V(ision))~\cite{lan2024translatotron} contains $512^2$ synthetic text images built from IWSLT14/17. Following the official splits, we evaluate De$\rightarrow$En (3{,}526), En$\rightarrow$De (3{,}524), Fr$\rightarrow$En (4{,}925), and Ro$\rightarrow$En (1{,}008).
\textbf{PRIM}~\cite{tian2025prim} contains photographed real-world $512\times46$ text strips. We evaluate all five directions En$\rightarrow$\{De, Fr, Cs, Ru, Ro\} (340 each) against the official Google-translation references.
\textbf{ScenePair}~\cite{mostel} provides 1{,}280 real scene photographs with English text. We translate every whole scene image into Chinese (En$\rightarrow$Zh). For a controlled comparison, all methods use the same text boxes and target translations, with each method receiving them in its official input format. Fixing these upstream OCR and translation stages prevents their errors from confounding the comparison. The resulting protocol isolates the rendering stage and measures how reliably each method places the prescribed Chinese text within the specified image regions. Exact-match ACC is computed by the recognizer of PP-OCRv6~\cite{zhang2026pp}.

\paragraph{Protocol.}
Following~\citet{unitranslator2026}, translated text is read back from the edited image with EasyOCR v1.7. We use per-direction language packs, sort boxes by top-left coordinate, and join them with spaces as in the official PRIM script. We score the text with sacreBLEU using the 13a tokenizer. TTV is lowercased following the IWSLT convention, while PRIM uses both official reference sets. We also report COMET with \texttt{wmt22-comet-da}~\cite{rei2020comet}. BLEU follows the standard machine-translation metric~\cite{papineni2002bleu}. For TTV, image quality is measured by SSIM against the source, using the structural-similarity formulation of~\citet{wang2004ssim}.
To reduce sensitivity to initialization, we evaluate our configurations with multiple random seeds and report the mean across seeds. For each seed, the Avg. columns are then computed as unweighted means across translation directions.
OCR read-back systematically underestimates rendered-text quality, and reader errors are attributed to the system under test. We report EasyOCR numbers in all comparisons for protocol parity.
An LLM-bridged alignment check also found $\approx$1.4\% of TTV De$\leftrightarrow$En reference lines locally misaligned by $\pm 1$ line, a shared defect of the reference files that shifts corpus BLEU on our outputs by well under one point.

\paragraph{Implementation.}
We use Z-Image-Turbo~\cite{zimage}, a distilled 6B flow-matching T2I model, with $T{=}50$ steps and editing window $n_{\max}{=}45$.
For all TTV and PRIM runs, we set RCR $\kappa{=}0.5$ ($\tau_{\mathrm{hi}}{=}0.6$, $\tau_{\mathrm{lo}}{=}0.3$), CE $c{=}0.4$, velocity gain $\gamma{=}1.3$, and latent-mask dilation to 15\%.
Images are resized so the long side is 1024 using bicubic interpolation, with dimensions rounded to multiples of 16. PRIM strips are bicubic-upscaled to the same editing canvas ($512\times46 \to 1024\times80$) and downscaled to their original size after editing.
Text regions and recognized strings come from the PP-OCRv5~\cite{ppocr}.
Translations and captions come from one call per image to a commercial LLM (Doubao-Seed-2.0-lite). The LLM returns the translations embedded in a pair of descriptive captions. The source caption $c_{\mathrm{src}}$ describes the scene with the original strings in quotes. The target caption $c_{\mathrm{tgt}}$ is identical except that the quoted strings are replaced by their translations.
The prompt contains the image and the numbered OCR text lines. It asks the LLM to translate each line into a concise target-language scene-text phrase and to return strict JSON with the translation pairs, a one-sentence English source caption that quotes every source line verbatim, and a target caption that changes only those quoted strings to their translations. The prompt also requires the original line order and all non-text scene descriptions to remain unchanged.
All experiments are conducted on a single NVIDIA RTX 4090 GPU.

\begin{figure}[!htbp]
\centering
\includegraphics[width=0.45\textwidth]{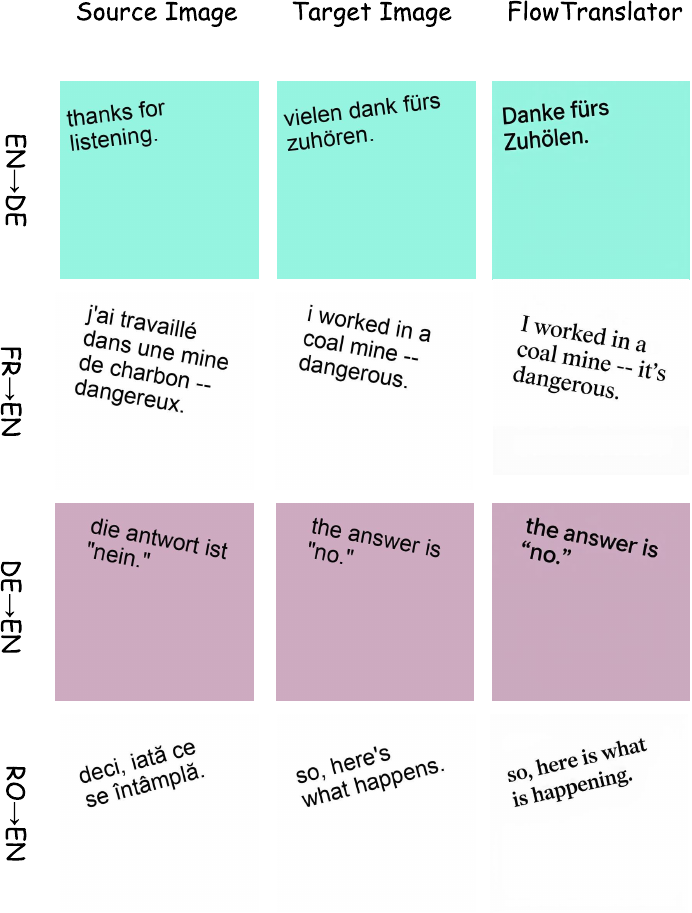}
\caption{Qualitative results on TTV. Each row shows the source image, the benchmark target image, and the FlowTranslator output for one translation direction. The examples cover EN$\rightarrow$DE, FR$\rightarrow$EN, DE$\rightarrow$EN, and RO$\rightarrow$EN.}
\label{fig:ttv-qualitative}
\end{figure}

\begin{table*}[t]
\centering
\caption{PRIM benchmark with EasyOCR read-back. BLEU uses both official reference sets, while COMET uses the Google references. Averages are computed from unrounded per-direction scores.}
\label{tab:prim}
{\small
\setlength{\tabcolsep}{3.6pt}
\begin{tabular}{l c ccccc cc}
\toprule
& & \multicolumn{5}{c}{BLEU$\uparrow$ / COMET$\uparrow$} & \multicolumn{2}{c}{Avg.} \\
\cmidrule(lr){3-7}\cmidrule(lr){8-9}
Model & IIMT-tuned & En-De & En-Fr & En-Cs & En-Ru & En-Ro & BLEU & COMET \\
\midrule
PARSeq-mT-Render~\cite{tian2025prim} & \checkmark & 9.5\,/\,41.7 & 13.8\,/\,46.9 & 7.7\,/\,43.9 & 5.5\,/\,48.1 & 12.8\,/\,53.5 & 9.9 & 46.8 \\
PEIT-Render~\cite{zhu2023peit} & \checkmark & 10.4\,/\,45.1 & 14.0\,/\,48.1 & 7.9\,/\,46.2 & 5.3\,/\,47.7 & 14.2\,/\,52.9 & 10.4 & 48.0 \\
Translatotron-V~\cite{lan2024translatotron} & \checkmark & 1.7\,/\,34.3 & 1.9\,/\,30.2 & 1.1\,/\,30.5 & 0.9\,/\,32.0 & 1.3\,/\,33.9 & 1.4 & 32.2 \\
VisTrans~\cite{tian2025prim} & \checkmark & 12.6\,/\,44.4 & 17.0\,/\,49.4 & 5.9\,/\,41.8 & 7.2\,/\,49.4 & 13.9\,/\,50.2 & 11.3 & 47.0 \\
UniTranslator~\cite{unitranslator2026} & \checkmark & 13.1\,/\,48.7 & 21.6\,/\,58.5 & 7.4\,/\,43.3 & \textbf{8.4}\,/\,\textbf{50.2} & 13.5\,/\,52.8 & 12.8 & 50.7 \\
\midrule
Masked FlowEdit~\cite{kulikov2025flowedit} &  & 29.9\,/\,56.7 & 24.6\,/\,57.7 & 15.5\,/\,56.9 & 2.2\,/\,41.1 & 12.3\,/\,54.3 & 16.9 & 53.3 \\
FlowTranslator (Ours) &  & \textbf{35.2}\,/\,\textbf{63.0} & \textbf{28.5}\,/\,\textbf{60.3} & \textbf{19.8}\,/\,\textbf{61.1} & 4.6\,/\,43.9 & \textbf{14.9}\,/\,\textbf{57.4} & \textbf{20.6} & \textbf{57.2} \\
\bottomrule
\end{tabular}}

\end{table*}

\subsection{Main Results}

\paragraph{TTV.}
Tab.~\ref{tab:ttv} compares FlowTranslator with training-based and training-free IIMT systems.
Qwen-Image performs poorly with an average BLEU of 0.22, whereas GPT-Image-1 reaches a competitive average of 23.06.
FlowTranslator attains the best average BLEU, 23.19 versus 21.17 for UniTranslator, the strongest IIMT-tuned system. It exceeds UniTranslator on three of four directions: 25.15 versus 25.03 on De$\rightarrow$En, 17.15 versus 13.41 on En$\rightarrow$De, and 23.37 versus 18.45 on Ro$\rightarrow$En. On Fr$\rightarrow$En, it scores 27.07 versus 27.77.
Against our own masked-FlowEdit baseline with the same hyperparameters, the controls are roughly neutral on TTV's large clean glyphs ($-0.4$ to $+0.7$ BLEU).
En$\rightarrow$De remains our weakest direction at 17.15. A plausible, though unverified, contributor is reader fragility on umlauts and compounds.
Our average SSIM (0.746) is below the IIMT-tuned systems' range of 0.77 to 0.80. Stronger edits deviate more from the reference renders.
Fig.~\ref{fig:ttv-qualitative} provides representative examples from all four TTV directions. FlowTranslator renders the translated content with layouts and visual styles close to the target images while preserving the source background.

\paragraph{PRIM.}
On photographed real-world strips (Tab.~\ref{tab:prim}), FlowTranslator obtains the highest reported scores on every direction except En$\rightarrow$Ru. For example, it scores 35.2 / 63.0 on En$\rightarrow$De versus 13.1 / 48.7 for UniTranslator. It also achieves the highest average BLEU, 20.6 versus 12.8, and COMET, 57.2 versus 50.7, despite UniTranslator being fine-tuned on 7M in-domain pairs.
Over our own masked-FlowEdit baseline with the same hyperparameters, the controls add +2.4 to +5.3 BLEU per direction (average 16.9$\to$20.6, COMET 53.3$\to$57.2).
Our failure direction is En$\rightarrow$Ru. Z-Image-Turbo's prior renders Cyrillic unreliably, producing near-Latin glyph substitutions that OCR cannot read and resulting in 4.6 BLEU. Training-free operation inherits the script coverage of the frozen backbone.

\begin{table}[ht]
\centering
\caption{ScenePair En$\rightarrow$Zh whole-image translation on 1{,}280 real scenes. All methods share the same text boxes and target translations. ``STE-tr.'' marks scene-text-editing training and ``Zh'' native Chinese glyph support. Rows without ``Zh'' are capability diagnostics rather than head-to-head competitors.}
\label{tab:scenepair}
{\small
\setlength{\tabcolsep}{5pt}
\begin{tabular}{l c c c c}
\toprule
Model & STE-tr. & Zh & ACC$\uparrow$ & SSIM$\uparrow$ \\
\midrule
TextDiffuser~\cite{textdiffuser} & \checkmark &  & 0.00 & 0.9434 \\
DiffSTE~\cite{diffste} & \checkmark &  & 0.23 & \textbf{0.9796} \\
AnyText~\cite{anytext} & \checkmark & \checkmark & 36.80 & 0.8157 \\
TextFlux~\cite{textflux} & \checkmark & \checkmark & 14.30 & 0.9506 \\
FLUX.1-Fill~\cite{flux2024} &  &  & 0.39 & 0.9231 \\
FLUX.1-Kontext~\cite{kontext} &  &  & 0.00 & 0.4273 \\
\midrule
Masked FlowEdit~\cite{kulikov2025flowedit} &  & \checkmark & 33.44 & 0.9533 \\
FlowTranslator (Ours) &  & \checkmark & \textbf{40.08} & 0.9509 \\
\bottomrule
\end{tabular}}

\end{table}

\subsection{Real-Scene Generalization}

\begin{figure*}[t]
\centering
\includegraphics[width=0.95\textwidth]{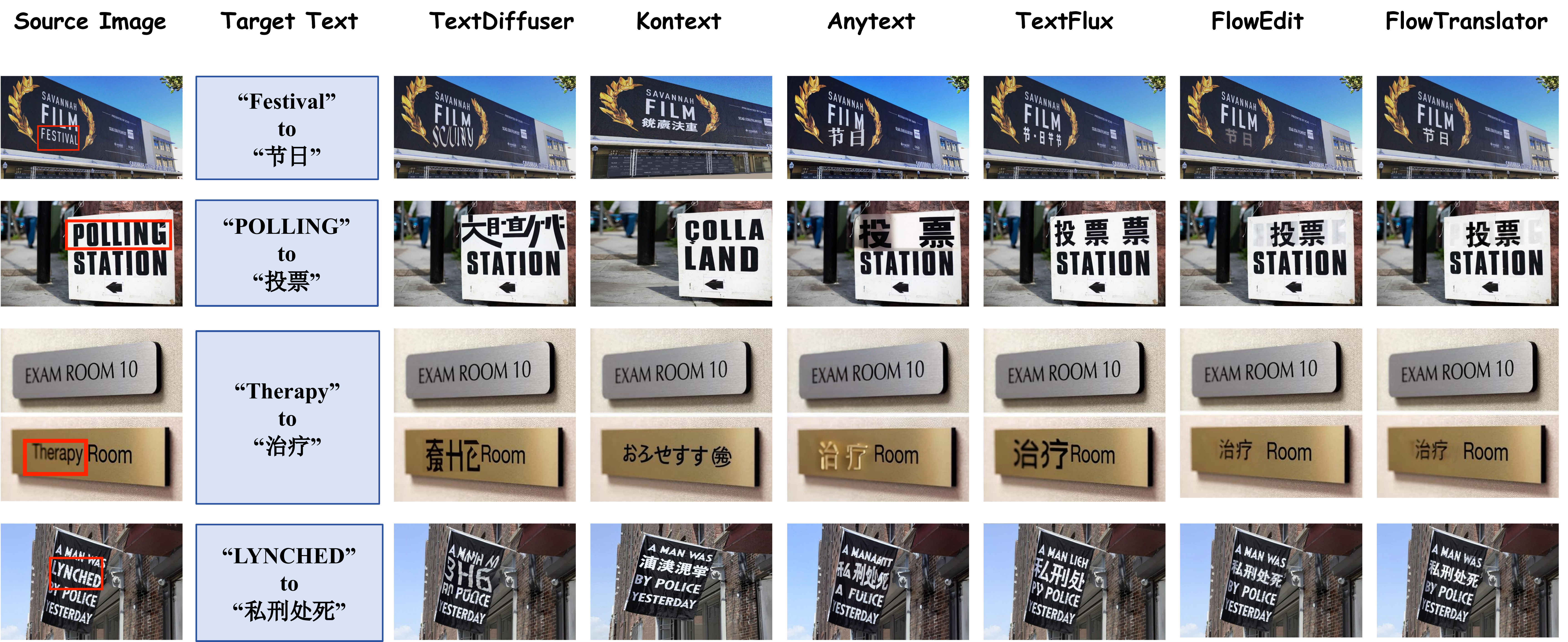}
\caption{Qualitative comparison on whole-image ScenePair En→Zh translation. For each example, we show the source image (with red boxes marking text to be translated), the requested target text, and outputs of existing open-source methods, followed by FlowTranslator. FlowTranslator renders the requested Chinese text within the original region. It preserves both the surrounding scene and the local text appearance.}
\label{fig:scenepair-comparison}
\end{figure*}

Under this rendering-focused protocol, Tab.~\ref{tab:scenepair} evaluates FlowTranslator against six existing editing systems.
FlowTranslator reaches 40.08\% exact-match ACC with an SSIM of 0.9509. Among the two trained editors with Chinese glyph support, it exceeds AnyText by 3.28 points and TextFlux by 25.78 points. The other four systems remain below 0.4\% exact-match ACC in this heterogeneous stress test. DiffSTE has the highest SSIM at 0.9796 but near-zero ACC, while Kontext has the lowest SSIM at 0.4273.
Against the Masked FlowEdit baseline in the same pipeline, the controls lift exact-match ACC from 33.44\% to 40.08\%, an increase of 6.64 points, while SSIM changes from 0.9533 to 0.9509
Fig.~\ref{fig:scenepair-comparison} shows qualitative outputs on ScenePair.  FlowTranslator replaces the selected English text with the requested Chinese translation while retaining the surrounding scene and the local appearance of the text region.

\subsection{Ablations}
\label{sec:ablations}

\paragraph{Trajectory Controls.}
Holding the pipeline, masks, prompts, and all other hyperparameters fixed, adding RCR and CE raises average PRIM BLEU from 16.88 to 20.57, a relative gain of 22\%. The two single-mechanism arms show where this gain comes from. CE alone already recovers 3.33 of the full 3.69-point improvement, roughly 90\% of the total. RCR is nonetheless not redundant: it composes positively with CE on four of the five directions, and only on En$\rightarrow$Ru does adding RCR on top of CE cost a small amount, lowering BLEU from 5.03 to 4.57. This division of labor matches the mechanism described in \S\ref{sec:method}: CE supplies the direction that drives most of the edit, while RCR supplies the room for that direction to act.

\begin{table}[ht]
\centering
\caption{Component ablations. Control rows are full PRIM runs (340 images per direction, averaged over the five directions) sharing the same hyperparameters ($\gamma{=}1.3$, 15\% dilation, RCR $\kappa{=}0.5$, CE $c{=}0.4$) and differing only in the mechanism switches.}
\label{tab:ablation}
{\small
\setlength{\tabcolsep}{3pt}
\begin{tabular}{l l c}
\toprule
Component & Setting & BLEU$\uparrow$ \\
\midrule
\multirow{4}{*}{Controls (PRIM avg.)} & Masked FlowEdit (Baseline) & 16.88 \\
 & +RCR only & 17.40 \\
 & +CE only & 20.21 \\
 & +RCR+CE (Ours) & \textbf{20.57} \\
\midrule
\multirow{3}{*}{Edit window $n_{\max}$} & 43 & 18.95 \\
 & 45 (Default) & \textbf{20.57} \\
 & 48 & 14.99 \\
\bottomrule
\end{tabular}}

\end{table}

\paragraph{Edit Window.}
The edit window $n_{\max}$ controls how many of the $T$ solver steps carry the edit. Tab.~\ref{tab:ablation} shows a clear peak at $n_{\max}{=}45$: shrinking it to 43 leaves visible source text and drops BLEU to 18.95, while extending it to 48 over-edits the region and destroys layout, dropping BLEU further to 14.99. We use $n_{\max}{=}45$ for $T{=}50$ in all other experiments as the setting that balances these two failure modes.
The trajectory controls and the edit window act on different benchmarks in a consistent direction. On ScenePair, adding the controls raises exact-match ACC from 33.44\% to 40.08\% (Tab.~\ref{tab:scenepair}), the same pattern of improvement seen on PRIM.

\paragraph{Cross-benchmark Analysis.}
Across benchmarks, the relative gain from the trajectory controls tracks how hard the edit is to ignite. The gain is largest on PRIM, where the translated text occupies a small, photographic strip of the image (+22\% relative average BLEU). But it is smallest on TTV, where the text is large, clean, and synthetic (+2\% on Ro$\rightarrow$En, its most affected direction). 
The benchmarks differ in more than text-area ratio, so this pattern is consistent with the ignition interpretation that hard-to-ignite, small-glyph regimes benefit most from explicit trajectory intervention.
Importantly, the same control configuration is used across benchmarks, so the difference in gains does not arise from benchmark-specific tuning.
The agreement between PRIM BLEU and ScenePair exact-match ACC further shows that the controls improve text completion under both strip-level and whole-image evaluation.
\section{Conclusion}

We presented FlowTranslator, a training-free IIMT framework. It replaced the trained rendering stage of an IIMT cascade with region-masked flow editing on a frozen text-to-image model. Two mask-confined trajectory controls, Region Coupling Release and Coupling Expansion, resolved an ignition failure specific to the text-minority regime of cross-lingual editing. 
Under the EasyOCR read-back protocol, FlowTranslator attained the highest reported average BLEU and COMET on PRIM. It also attained the highest average BLEU on Translatotron-V(ision), all without any IIMT training data. On real ScenePair photographs translated into Chinese, it reached 40.1\% exact-match accuracy, exceeding six released editing systems. Once its branch coupling was corrected for the text-minority regime, a frozen T2I flow model could match or exceed systems trained on millions of IIMT pairs. This established flow-based editing as a strong training-free baseline for IIMT.
The method still inherited the script coverage of its frozen backbone, and Cyrillic rendering failed in our experiments. Extending the trajectory controls to backbones with broader script support remains a direction for future work.

\bibliographystyle{plainnat}
\bibliography{references}

@inproceedings{mansimov2020towards,
  title={Towards End-to-End In-Image Neural Machine Translation},
  author={Mansimov, Elman and Stern, Mitchell and Chen, Mia and Firat, Orhan and Uszkoreit, Jakob and Jain, Puneet},
  booktitle={Proceedings of the First International Workshop on Natural Language Processing Beyond Text},
  pages={70--74},
  year={2020}
}

@misc{openai2025gptimage,
  title={Introducing Our Latest Image Generation Model in the API},
  author={{OpenAI}},
  howpublished={\url{https://openai.com/index/image-generation-api/}},
  month={April},
  year={2025}
}

@article{wu2025qwenimage,
  title={Qwen-Image Technical Report},
  author={Wu, Chenfei and Li, Jiahao and Zhou, Jingren and Lin, Junyang and Gao, Kaiyuan and Yan, Kun and Yin, Shengming and Bai, Shuai and Xu, Xiao and Chen, Yilei and Chen, Yuxiang and Tang, Zecheng and Zhang, Zekai and Wang, Zhengyi and Yang, An and Yu, Bowen and Cheng, Chen and Liu, Dayiheng and Li, Deqing and Zhang, Hang and Meng, Hao and Wei, Hu and Ni, Jingyuan and Chen, Kai and Cao, Kuan and Peng, Liang and Qu, Lin and Wu, Minggang and Wang, Peng and Yu, Shuting and Wen, Tingkun and Feng, Wensen and Xu, Xiaoxiao and Wang, Yi and Zhang, Yichang and Zhu, Yongqiang and Wu, Yujia and Cai, Yuxuan and Liu, Zenan},
  journal={arXiv preprint arXiv:2508.02324},
  year={2025}
}

@inproceedings{lan2024translatotron,
  title={Translatotron-V(ision): An End-to-End Model for In-Image Machine Translation},
  author={Lan, Zhibin and Niu, Liqiang and Meng, Fandong and Zhou, Jie and Zhang, Min and Su, Jinsong},
  booktitle={Findings of the Association for Computational Linguistics: ACL 2024},
  pages={5472--5485},
  year={2024}
}

@inproceedings{tian2025exploring,
  title={Exploring In-Image Machine Translation with Real-World Background},
  author={Tian, Yanzhi and Liu, Zeming and Liu, Zhengyang and Guo, Yuhang},
  booktitle={Findings of the Association for Computational Linguistics: ACL 2025},
  year={2025}
}

@article{tian2025prim,
  title={PRIM: Towards Practical In-Image Multilingual Machine Translation},
  author={Tian, Yanzhi and Liu, Zeming and Liu, Zhengyang and Feng, Chong and Li, Xin and Huang, Heyan and Guo, Yuhang},
  journal={arXiv preprint arXiv:2509.05146},
  year={2025}
}

@article{unitranslator2026,
  title={UniTranslator: A Unified Multi-modal Framework for End-to-end In-Image Machine Translation},
  author={Lyu, Jiahao and Fu, Pei and Li, Zhenhang and Zhang, Shaojie and Yang, Jiahui and Zhou, Yu and Ma, Can and Luo, Zhenbo and Luan, Jian},
  journal={arXiv preprint arXiv:2606.24333},
  year={2026}
}

@inproceedings{zhu2023peit,
  title={{PEIT}: Bridging the Modality Gap with Pre-trained Models for End-to-End Image Translation},
  author={Zhu, Shaolin and Li, Shangjie and Lei, Yikun and Xiong, Deyi},
  booktitle={Proceedings of the 61st Annual Meeting of the Association for Computational Linguistics},
  pages={13433--13447},
  year={2023}
}

@inproceedings{lan2023exploring,
  title={Exploring Better Text Image Translation with Multimodal Codebook},
  author={Lan, Zhibin and Yu, Jiawei and Li, Xiang and Zhang, Wen and Luan, Jian and Wang, Bin and Huang, Degen and Su, Jinsong},
  booktitle={Proceedings of the 61st Annual Meeting of the Association for Computational Linguistics},
  pages={3479--3491},
  year={2023}
}

@inproceedings{kulikov2025flowedit,
  title={FlowEdit: Inversion-Free Text-Based Editing Using Pre-Trained Flow Models},
  author={Kulikov, Vladimir and Kleiner, Matan and Huberman-Spiegelglas, Inbar and Michaeli, Tomer},
  booktitle={Proceedings of the IEEE/CVF International Conference on Computer Vision},
  pages={19721--19730},
  year={2025}
}

@inproceedings{textflow2026,
  title={Towards Training-Free Scene Text Editing},
  author={Li, Yubo and Qin, Xugong and Zhang, Peng and Lin, Hailun and Zeng, Gangyan and Zhang, Kexin},
  booktitle={Proceedings of the IEEE/CVF Conference on Computer Vision and Pattern Recognition},
  month={June},
  year={2026},
  pages={15291--15301}
}

@inproceedings{ye2025textssr,
  title={TextSSR: diffusion-based data synthesis for scene text recognition},
  author={Ye, Xingsong and Du, Yongkun and Tao, Yunbo and Chen, Zhineng},
  booktitle={2025 IEEE/CVF International Conference on Computer Vision (ICCV)},
  pages={17464--17473},
  year={2025},
  organization={IEEE}
}

@inproceedings{textriro,
  title={Real In, Real Out: What If We Only Use Real Data for Scene Text Editing?},
  author={Ye, Xingsong and Du, Yongkun and Zhang, Jiaxin and Sun, Chong and Chen, Zhineng},
  booktitle={Advances in Neural Information Processing Systems},
  year={2026}
}

@inproceedings{mostel,
  title={Exploring Stroke-Level Modifications for Scene Text Editing},
  author={Qu, Yadong and Tan, Qingfeng and Xie, Hongtao and Xu, Jianjun and Wang, Yuxin and Zhang, Yongdong},
  booktitle={Proceedings of the AAAI Conference on Artificial Intelligence},
  volume={37},
  pages={2119--2127},
  year={2023}
}

@inproceedings{anytext,
  title={AnyText: Multilingual Visual Text Generation and Editing},
  author={Tuo, Yuxiang and Xiang, Wangmeng and He, Jun-Yan and Geng, Yifeng and Xie, Xuansong},
  booktitle={International Conference on Learning Representations},
  year={2024}
}

@inproceedings{esser2024sd3,
  title={Scaling Rectified Flow Transformers for High-Resolution Image Synthesis},
  author={Esser, Patrick and Kulal, Sumith and Blattmann, Andreas and Entezari, Rahim and M{\"u}ller, Jonas and Saini, Harry and Levi, Yam and Lorenz, Dominik and Sauer, Axel and Boesel, Frederic and Podell, Dustin and Dockhorn, Tim and English, Zion and Lacey, Kyle and Goodwin, Alex and Marek, Yannik and Rombach, Robin},
  booktitle={International Conference on Machine Learning},
  year={2024}
}

@misc{flux2024,
  title={{FLUX}},
  author={{Black Forest Labs}},
  howpublished={\url{https://github.com/black-forest-labs/flux}},
  year={2024}
}

@article{flowalign,
  title={FlowAlign: Trajectory-Regularized, Inversion-Free Flow-based Image Editing},
  author={Kim, Jeongsol and Hong, Yeobin and Park, Jonghyun and Ye, Jong Chul},
  journal={arXiv preprint arXiv:2505.23145},
  year={2025}
}

@article{lipman2022flowmatching,
  title={Flow Matching for Generative Modeling},
  author={Lipman, Yaron and Chen, Ricky T. Q. and Ben-Hamu, Heli and Nickel, Maximilian and Le, Matt},
  journal={arXiv preprint arXiv:2210.02747},
  year={2022}
}

@article{fang2025rsste,
  title={Recognition-Synergistic Scene Text Editing},
  author={Fang, Zhengyao and Lyu, Pengyuan and Wu, Jingjing and Zhang, Chengquan and Yu, Jun and Lu, Guangming and Pei, Wenjie},
  journal={arXiv preprint arXiv:2503.08387},
  year={2025}
}

@article{zimage,
  title={Z-Image: An Efficient Image Generation Foundation Model with Single-Stream Diffusion Transformer},
  author={{Z-Image Team}},
  journal={arXiv preprint arXiv:2511.22699},
  year={2025}
}

@article{ppocr,
  title={PaddleOCR 3.0 Technical Report},
  author={Cui, Cheng and Sun, Ting and Lyu, Manhui and Liu, Tingquan and Zhang, Zelun and Ma, Yikang and Ma, Shuangqing and Yu, Zhihao and Song, Zhipeng and Guo, Jinhang and others},
  journal={arXiv preprint arXiv:2507.05595},
  year={2025}
}

@inproceedings{du2025svtrv2,
  title={Svtrv2: Ctc beats encoder-decoder models in scene text recognition},
  author={Du, Yongkun and Chen, Zhineng and Xie, Hongtao and Jia, Caiyan and Jiang, Yu-Gang},
  booktitle={2025 IEEE/CVF International Conference on Computer Vision (ICCV)},
  pages={20147--20156},
  year={2025},
  organization={IEEE}
}

@inproceedings{du2026unirec,
  title={Unirec-0.1 b: Unified text and formula recognition with 0.1 b parameters},
  author={Du, Yongkun and Chen, Zhineng and Xie, Yazhen and Bai, Weikang and Feng, Hao and Shi, Wei and Su, Yuchen and Huang, Can and Jiang, Yu-Gang},
  booktitle={European Conference on Computer Vision},
  pages={36--55},
  year={2026},
  organization={Springer}
}

@article{zhang2026pp,
  title={PP-OCRv6: From 1.5 M to 34.5 M Parameters, Surpassing Billion-Scale VLMs on OCR Tasks},
  author={Zhang, Yubo and Wang, Xueqing and Lin, Manhui and Zhang, Yue and Deng, Penglongyi and Sun, Ting and Gao, Tingquan and Zhang, Zelun and Liu, Jiaxuan and Zhou, Changda and others},
  journal={arXiv preprint arXiv:2606.13108},
  year={2026}
}

@inproceedings{textdiffuser,
  title={TextDiffuser: Diffusion Models as Text Painters},
  author={Chen, Jingye and Huang, Yupan and Lv, Tengchao and Cui, Lei and Chen, Qifeng and Wei, Furu},
  booktitle={Advances in Neural Information Processing Systems},
  year={2023}
}

@article{diffste,
  title={Improving Diffusion Models for Scene Text Editing with Dual Encoders},
  author={Ji, Jiabao and Zhang, Guanhua and Wang, Zhaowen and Hou, Bairu and Zhang, Zhifei and Price, Brian and Chang, Shiyu},
  journal={Transactions on Machine Learning Research},
  year={2023}
}

@article{textflux,
  title={TextFlux: An OCR-Free DiT Model for High-Fidelity Multilingual Scene Text Synthesis},
  author={Xie, Yu and Zhang, Jielei and Chen, Pengyu and Wang, Ziyue and Wang, Weihang and Gao, Longwen and Li, Peiyi and Sun, Huyang and Zhang, Qiang and Qiao, Qian and Fan, Jiaqing and Lian, Zhouhui},
  journal={arXiv preprint arXiv:2505.17778},
  year={2025}
}

@article{kontext,
  title={FLUX.1 Kontext: Flow Matching for In-Context Image Generation and Editing in Latent Space},
  author={Batifol, Stephen and Blattmann, Andreas and Boesel, Frederic and Consul, Saksham and Diagne, Cyril and Dockhorn, Tim and English, Jack and English, Zion and Esser, Patrick and Kulal, Sumith and others},
  journal={arXiv preprint arXiv:2506.15742},
  year={2025}
}

@inproceedings{rei2020comet,
  title={{COMET}: A Neural Framework for {MT} Evaluation},
  author={Rei, Ricardo and Stewart, Craig and Farinha, Ana C and Lavie, Alon},
  booktitle={Proceedings of the 2020 Conference on Empirical Methods in Natural Language Processing},
  pages={2685--2702},
  year={2020}
}

@inproceedings{wu2019editingwild,
  title={Editing Text in the Wild},
  author={Wu, Liheng and Zhang, Cheng and Liu, Jun and Han, Jing and Liu, Jun and Ding, Errui and Bai, Xiang},
  booktitle={Proceedings of the 27th ACM International Conference on Multimedia},
  pages={1500--1508},
  year={2019}
}

@inproceedings{roy2020stefann,
  title={{STEFANN}: Scene Text Editor Using Font Adaptive Neural Network},
  author={Roy, Priyanka and Bhattacharya, Saumik and Ghosh, Subhankar and Pal, Umapada},
  booktitle={Proceedings of the IEEE/CVF Conference on Computer Vision and Pattern Recognition},
  pages={13228--13237},
  year={2020}
}

@inproceedings{yang2020swaptext,
  title={SwapText: Image Based Texts Transfer in Scenes},
  author={Yang, Qiang and Huang, Jing and Lin, Zhi},
  booktitle={Proceedings of the IEEE/CVF Conference on Computer Vision and Pattern Recognition},
  pages={14700--14709},
  year={2020}
}

@inproceedings{zhao2024udifftext,
  title={UDiffText: A Unified Framework for High-Quality Text Synthesis in Arbitrary Images via Character-Aware Diffusion Models},
  author={Zhao, Yiming and Lian, Zhouhui},
  booktitle={European Conference on Computer Vision},
  pages={217--233},
  year={2024}
}

@article{zeng2024textctrl,
  title={TextCtrl: Diffusion-Based Scene Text Editing with Prior Guidance Control},
  author={Zeng, Weichao and Shu, Yan and Li, Zhenhang and Yang, Dongbao and Zhou, Yu},
  journal={Advances in Neural Information Processing Systems},
  volume={37},
  pages={138569--138594},
  year={2024}
}

@inproceedings{qian2024anytrans,
  title={AnyTrans: Translate AnyText in the Image with Large Scale Models},
  author={Qian, Zhaoyang and Zhang, Peng and Yang, Bowen and Fan, Kai and Ma, Yiming and Wong, Derek F. and Sun, Xiang and Ji, Rongrong},
  booktitle={Findings of the Association for Computational Linguistics: EMNLP 2024},
  pages={2432--2444},
  year={2024}
}

@article{lan2025fluxtext,
  title={FLUX-Text: A Simple and Advanced Diffusion Transformer Baseline for Scene Text Editing},
  author={Lan, Rui and Bai, Yancheng and Duan, Xu and Li, Mingxing and Jin, Dongyang and Xu, Ryan and Nie, Dong and Sun, Lei and Chu, Xiangxiang},
  journal={arXiv preprint arXiv:2505.03329},
  year={2025}
}

@article{wang2024tamingrf,
  title={Taming Rectified Flow for Inversion and Editing},
  author={Wang, Jiangshan and Pu, Junfu and Qi, Zhongang and Guo, Jiayi and Ma, Yue and Huang, Nisha and Chen, Yuxin and Li, Xiu and Shan, Ying},
  journal={arXiv preprint arXiv:2411.04746},
  year={2024}
}

@article{zheng2025lanpaint,
  title={LanPaint: Training-Free Diffusion Inpainting with Asymptotically Exact and Fast Conditional Sampling},
  author={Zheng, Chen and Lan, Yuan and Wang, Yang},
  journal={arXiv preprint arXiv:2502.03491},
  year={2025}
}

@article{zhu2025kvedit,
  title={KV-Edit: Training-Free Image Editing for Precise Background Preservation},
  author={Zhu, Tianrui and Zhang, Shiyi and Shao, Jiawei and Tang, Yansong},
  journal={arXiv preprint arXiv:2502.17363},
  year={2025}
}

@inproceedings{papineni2002bleu,
  title={BLEU: A Method for Automatic Evaluation of Machine Translation},
  author={Papineni, Kishore and Roukos, Salim and Ward, Todd and Zhu, Wei-Jing},
  booktitle={Proceedings of the 40th Annual Meeting of the Association for Computational Linguistics},
  pages={311--318},
  year={2002}
}

@article{wang2004ssim,
  title={Image Quality Assessment: From Error Visibility to Structural Similarity},
  author={Wang, Zhou and Bovik, Alan C. and Sheikh, Hamid R. and Simoncelli, Eero P.},
  journal={IEEE Transactions on Image Processing},
  volume={13},
  number={4},
  pages={600--612},
  year={2004}
}

@article{xie2026decompose,
  title={Decompose Radicals, Then Reward: Fine-Grained Inspection for Accurate Chinese Text Rendering},
  author={Xie, Yazhen and Ye, Xingsong and Chen, Zhineng},
  journal={arXiv preprint arXiv:2609.37569},
  year={2026}
}

@inproceedings{du2022svtr,
  title     = {SVTR: Scene Text Recognition with a Single Visual Model},
  author = {Du, Yongkun and Chen, Zhineng and Jia, Caiyan and Yin, Xiaoting and Zheng, Tianlun and Li, Chenxia and Du, Yuning and Jiang, Yu-Gang},
  booktitle = {IJCAI},
  pages     = {884--890},
  year      = {2022},
}

@article{zheng2024cdistnet,
  title={Cdistnet: Perceiving multi-domain character distance for robust text recognition},
  author={Zheng, Tianlun and Chen, Zhineng and Fang, Shancheng and Xie, Hongtao and Jiang, Yu-Gang},
  journal={International Journal of Computer Vision},
  volume={132},
  number={2},
  pages={300--318},
  year={2024},
  publisher={Springer}
}

@article{ye2026all,
  title={All-in-One Multilingual Scene Text Recognition with Script-aware Mixture-of-Experts},
  author={Ye, Xingsong and Du, Yongkun and Zhang, Jiaxin and Li, Zhixian and Sun, Chong and Li, Chen and Lyu, Jing and Jin, Lianwen and Chen, Zhineng},
  journal={arXiv preprint arXiv:2609.24058},
  year={2026}
}

@inproceedings{ye2026advancing,
  title={Advancing WordArt-Oriented Scene Text Recognition: Datasets and Methods},
  author={Ye, Xingsong and Du, Yongkun and Zhang, Jiaxin and Zhang, Haojie and Sun, Chong and Li, Chen and Lyu, Jing and Chen, Zhineng},
  booktitle={European Conference on Computer Vision},
  pages={77--94},
  year={2026},
  organization={Springer}
}

@inproceedings{ye2026what,
  title={What's Wrong with Synthetic Data for Scene Text Recognition? A Strong Synthetic Engine with Diverse Simulations and Self-Evolution},
  author={Ye, Xingsong and Du, Yongkun and Zhang, JiaXin and Li, Chen and LYU, Jing and Chen, Zhineng},
  booktitle={2026 IEEE/CVF Conference on Computer Vision and Pattern Recognition (CVPR)},
  pages={16645-16654},
  year={2026},
  organization={IEEE}
}

\clearpage
\section*{Appendix}
\begingroup
\newcommand{\Min}{M}
\newcommand{\rin}{r_{\mathrm{in}}}
\newcommand{\rout}{r_{\mathrm{out}}}
\newcommand{\cosin}{\cos_{\mathrm{in}}}
\newcommand{\din}{d_{\mathrm{in}}}
\newcommand{\epstar}{\epsilon^{\mathrm{tar}}}

\setcounter{subsection}{0}
\setcounter{figure}{0}
\setcounter{table}{0}
\setcounter{equation}{0}
\renewcommand{\thesubsection}{A.\arabic{subsection}}
\renewcommand{\theHsubsection}{appendix.\arabic{subsection}}
\renewcommand{\thefigure}{S\arabic{figure}}
\renewcommand{\thetable}{S\arabic{table}}
\renewcommand{\theequation}{S\arabic{equation}}
\renewcommand{\theHfigure}{supp.\arabic{figure}}
\renewcommand{\theHtable}{supp.\arabic{table}}
\renewcommand{\theHequation}{supp.\arabic{equation}}
\raggedbottom

This appendix provides trajectory-level diagnostics for the two mask-confined
controls introduced in the main text. Using four matched arms (Masked
FlowEdit (denoted as FlowEdit in the figures), $+$RCR, $+$CE, and
$+$RCR$+$CE), we test whether RCR relaxes early branch coupling and whether CE
amplifies the accumulated displacement without destabilizing the latent
trajectory. All diagnostics use 12 paired images under the default setting
($T=50$, $n_{\max}=45$, $\kappa=0.5$, and $c=0.4$).

\subsection{Trajectory Diagnostics for Region Coupling Release}
\label{supp:sec:rcr}

\begin{figure}[!htbp]
\centering
\includegraphics[width=0.62\textwidth]{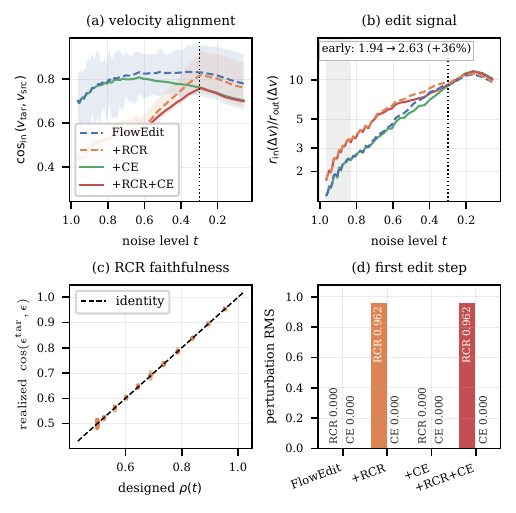}
\caption{RCR diagnostics: dashed/solid pairs nearly overlap at plotting
resolution (Masked FlowEdit with $+$CE, and $+$RCR with $+$RCR$+$CE).
(a) In-mask velocity alignment. Bands show the min--max range over 12 images
for Masked FlowEdit and $+$RCR. Dotted line: $\tau_{\mathrm{lo}}=0.3$.
(b) In-mask edit-signal concentration on a log scale. The first third of the
edit window is shaded ($+36\%$). (c) Prescribed versus realized noise
correlation (mean absolute error $0.0032$). (d) First-step RCR and CE
perturbations by arm. Zeros are explicitly labeled.}
\label{supp:fig:rcr}
\end{figure}

\begin{table*}[t]
\centering
\caption{Trajectory diagnostics over 12 images (45 edit steps each). Rows are
means across images except the two ``max'' rows, which are maxima over images and
steps. ``Early'' denotes the first third of the edit window. The ignition step
uses a single absolute threshold shared by all arms. Dashes mark probes that are
inactive in an arm by construction.}
\label{supp:tab:diag}
\small
\begin{tabular}{lcccc}
\toprule
Diagnostic & Masked FlowEdit & +RCR & +CE & +RCR+CE \\
\midrule
early $\cos_{\rm in}$ & 0.740 & 0.481 & 0.738 & 0.480 \\
early signal ratio & 1.94 & 2.63 & 1.93 & 2.63 \\
ignition step $\downarrow$ & 9.50 & 7.75 & 9.42 & 7.67 \\
final $d_{\rm in}$ & 0.440 & 0.443 & 0.533 & 0.513 \\
max rel.\ CE pert. & 0.000 & 0.000 & 0.267 & 0.246 \\
max $|Z_t^{\rm FE}|$ & 3.991 & 4.052 & 3.991 & 4.053 \\
$|\rho-\cos(\epsilon^{\rm tar},\epsilon)|$ & -- & 0.0032 & -- & 0.0032 \\
RCR pert.\ at first step & -- & 0.962 & -- & 0.962 \\
\bottomrule
\end{tabular}

\end{table*}

\paragraph{Paired diagnostic protocol.}
We instrument the sampling loop with per-step probes and compare four arms on
matched stochastic trajectories: Masked FlowEdit, $+\mathrm{RCR}$,
$+\mathrm{CE}$, and $+\mathrm{RCR}+\mathrm{CE}$. The arms differ only in
$(\kappa,c)$, set respectively to
$(0,0)$, $(0.5,0)$, $(0,0.4)$, and $(0.5,0.4)$, with
$\tau_{\mathrm{hi}}=0.6$ and $\tau_{\mathrm{lo}}=0.3$. Each arm uses the same 12
images, boxes, prompts, and seed for 50 scheduler steps, of which 45 are edit
steps, giving 540 per-step records per arm.
Let $\Min\in\{0,1\}^{1\times H\times W}$ be the binary region mask
broadcast over the $C$ latent channels, $Z^{\mathrm{src}}_0$ the clean source
latent, and $Z^{\mathrm{FE}}_t$ the running edit latent at noise level $t$. All
probes use mask-normalized root-mean-square (RMS) values, making them comparable
across region sizes:
$\rin(x)=\big((\sum \Min\odot x^{2})/(C\sum \Min)\big)^{1/2}$, and $\rout$
likewise over $1-\Min$. Let
$v_{\mathrm{src},t}=v_\theta(\hat z^{\mathrm{src}}_t,t,c_{\mathrm{src}})$ and
$v_{\mathrm{tar},t}=v_\theta(\hat z^{\mathrm{tar}}_t,t,c_{\mathrm{tgt}})$, and
let $\Delta v_t=v_{\mathrm{tar},t}-v_{\mathrm{src},t}$. To diagnose RCR, we log
the in-mask velocity alignment
$\cosin(v_{\mathrm{tar},t},v_{\mathrm{src},t})$ and the edit-signal
concentration $\rin(\Delta v_t)/\rout(\Delta v_t)$. RCR replaces the
target-branch noise inside the mask with a $\rho$-correlated sample,
$\epstar=\rho\,\epsilon+\sqrt{1-\rho^{2}}\,\epsilon'$ with
$\rho(t)=1-\kappa\,s(t)$ and $s$ a linear ramp that is $1$ for
$t\ge \tau_{\mathrm{hi}}$ and $0$ for $t\le \tau_{\mathrm{lo}}$, so it injects
$p^{\mathrm{RCR}}_t=\Min\odot t\,(\epstar-\epsilon)$. We log both the prescribed
$\rho(t)$ and the realized $\cos(\epstar,\epsilon)$. The probes require no
additional network evaluations and draw no random numbers, so enabling them
does not perturb the trajectory being measured.

\paragraph{Paired-trajectory control.}
Drawing RCR's extra noise $\epsilon'$ from the global generator would advance
the random stream and desynchronize every subsequent $\epsilon$, so the arms
would follow different trajectories and no per-step difference could be
attributed to the mechanism. We therefore draw $\epsilon'$ from a dedicated
fixed-seed generator, and we verify the result rather than assume it: at the
first edit step, where CE is inactive because the accumulated displacement is
zero, the Masked FlowEdit and $+$CE arms agree to the last logged digit on
\emph{all}
12 images, while $+$RCR differs on all 12. The update is also region-masked, so
$\rout(Z^{\mathrm{FE}}_t-Z^{\mathrm{src}}_0)=0$ exactly at every step of every
arm. With $n=12$ paired images, a unanimous direction in a sign test corresponds
to a two-sided $p=4.9\times10^{-4}$. All reported $p$-values are exact two-sided
paired sign tests with ties excluded.

\paragraph{In-mask edit-signal concentration.}
RCR lowers the early in-mask alignment of the two velocity fields from
$\cosin=0.740$ to $0.481$, so the residual $\Delta v_t$ that drives the edit is
subject to less cancellation inside the region. The concentration of that
residual in the mask rises from $1.94$ to $2.63$, i.e.\ $+36\%$
(Fig.~\ref{supp:fig:rcr}(a,b)). Both effects hold on $12/12$ images
($p=4.9\times10^{-4}$ each), and both originate at high noise: RCR stops
injecting additional decorrelation once the schedule reaches
$\tau_{\mathrm{lo}}=0.3$, after which late glyph refinement proceeds under
restored branch-noise coupling. The comparison also
localizes the cause: $+$CE alone moves the alignment by only $-0.002$, so it
is the branch-noise decorrelation, not the displacement amplification, that
produces the stronger in-mask signal.

\paragraph{Realized noise coupling.}
The mean absolute difference between prescribed $\rho(t)$ and measured
$\cos(\epstar,\epsilon)$ is $0.0032$ across all steps and images, including the
linear ramp (Fig.~\ref{supp:fig:rcr}(c)). This agreement confirms that the
implemented correlation follows the prescribed schedule.

\paragraph{First-step activation.}
CE injects $\alpha(t)\Min\odot\delta_t$, which is identically zero while
$\delta_t=0$. RCR injects $\Min\odot t(\epstar-\epsilon)$ independently of the
accumulated displacement and is therefore active at the noisy end: its
first-step magnitude averages $0.962$ over 12 images, versus $0.000$ for CE
(Fig.~\ref{supp:fig:rcr}(d)). The two terms are thus active in complementary
phases, RCR decaying as the branches recouple and CE peaking in mid-trajectory
under its $\sin^{2}(\pi t)$ gate.

\subsection{Trajectory Diagnostics for Coupling Expansion}
\label{supp:sec:ce}

\paragraph{Paired displacement probes.}
Using the same four paired arms, we measure the in-mask displacement from the
source,
\begin{equation}
\din(t)=\rin\!\left(Z^{\mathrm{FE}}_t-Z^{\mathrm{src}}_0\right).
\label{supp:eq:disp}
\end{equation}
CE amplifies the accumulated displacement
$\delta_t=Z^{\mathrm{FE}}_t-Z^{\mathrm{src}}_0$ by
$\alpha(t)=c\sin^{2}(\pi t)$, injecting
$p^{\mathrm{CE}}_t=\alpha(t)\,\Min\odot\delta_t$. We log this perturbation
relative to the target-branch latent it modifies, together with the per-step
update magnitude and $\max|Z^{\mathrm{FE}}_t|$. The update is region-masked, and
we verify
$\rout(Z^{\mathrm{FE}}_t-Z^{\mathrm{src}}_0)=0$ at every step of every arm.

\begin{figure}[!htbp]
\centering
\includegraphics[width=0.62\textwidth]{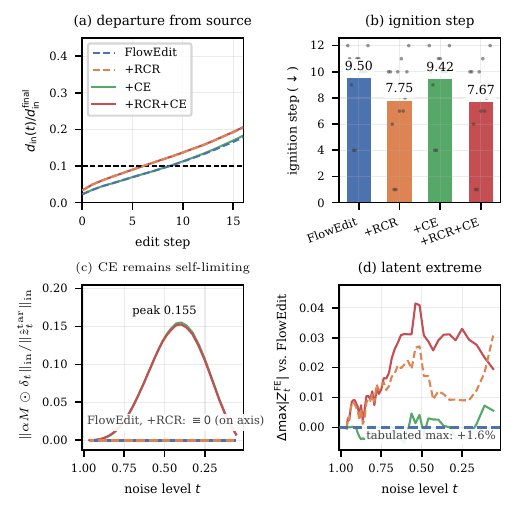}
\caption{CE effectiveness and stability: pairing as in Fig.~\ref{supp:fig:rcr}.
(a) In-mask displacement normalized by the Masked FlowEdit final value. Dashed
line: shared 10\% ignition threshold. (b) Ignition step with per-image values.
(c) Relative CE perturbation. CE-off arms lie at zero. (d) Paired per-step
difference in the latent maximum relative to Masked FlowEdit. Global maxima for
Masked FlowEdit and $+$RCR$+$CE differ by $1.6\%$.}
\label{supp:fig:ce}
\end{figure}

\begin{figure*}[!t]
\centering
\includegraphics[width=0.92\textwidth]{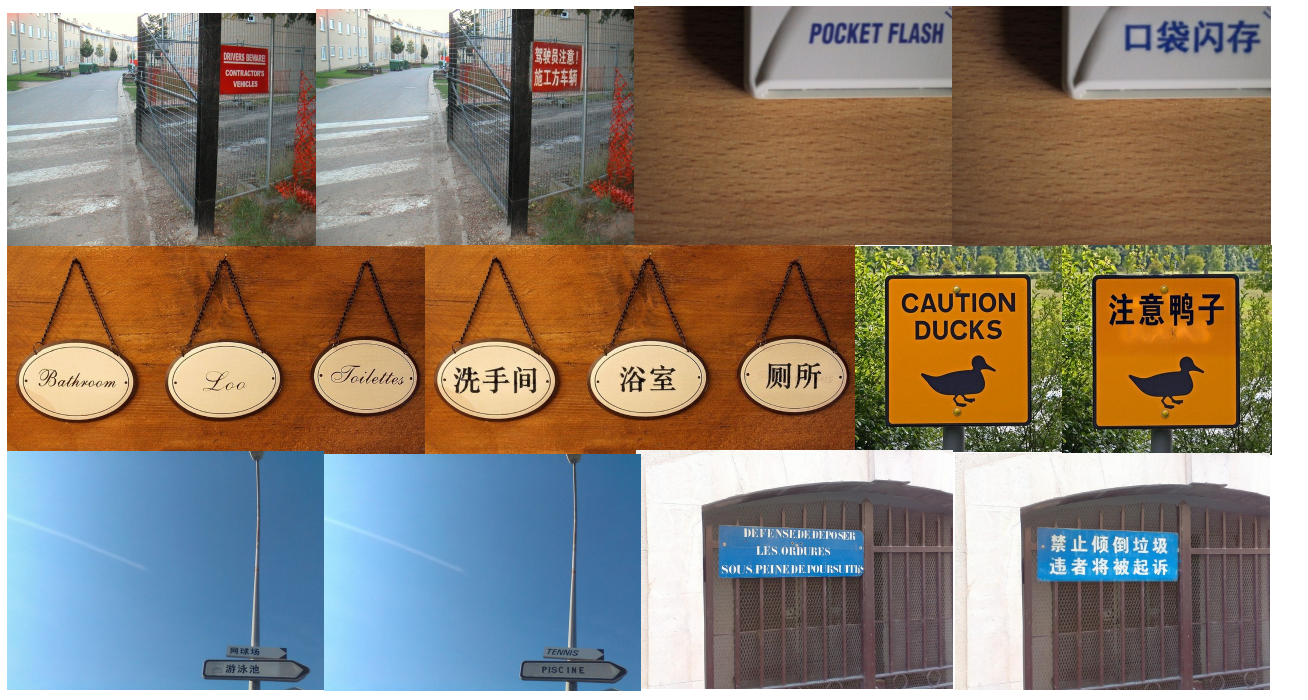}
\caption{Additional qualitative examples of full-image scene text translation
into Chinese. Each pair shows the source image on the left and the corresponding
FlowTranslator output on the right.}
\label{supp:fig:additional-qualitative}
\end{figure*}

\paragraph{Final edit displacement.}
CE raises the final in-mask displacement from $0.440$ to $0.533$, i.e.\ $+21\%$,
on $11/12$ images ($p=6.3\times10^{-3}$), and to $0.513$ ($+16\%$) alongside RCR
on $10/12$ images ($p=0.039$). This division of labor matches the mechanism:
RCR alone changes the final displacement by only $+0.7\%$ and primarily affects
edit onset, whereas CE affects final magnitude. Using an ignition threshold
shared by all arms (the first step at which $\din(t)$ reaches $10\%$ of the
Masked FlowEdit arm's final displacement on the same image), the edit departs
from the source at step $9.50$ for the baseline and $7.75$ under RCR, i.e.\
$1.75$ steps earlier,
earlier on $9/12$ images, later on $1/12$ and tied on $2/12$ ($p=0.021$). $+$CE
alone reaches $9.42$ while $+$RCR$+$CE reaches $7.67$, earlier on $10/12$
($p=0.012$) (Fig.~\ref{supp:fig:ce}(a,b)). In this controlled comparison, the
earlier ignition is therefore attributable to RCR, consistent with CE having
nothing to amplify at the start.

\paragraph{Self-limiting feedback.}
Because CE amplifies the displacement it helps create, we measure the resulting
perturbation. Its magnitude relative to the target-branch latent follows the
intended $\sin^{2}(\pi t)$ profile, rising to a mean peak of $0.155$ near
$t\!\approx\!0.5$ and returning to nearly zero before glyph refinement
(largest value over all images and steps $0.267$). The per-step update grows
smoothly rather than spiking. The latent maximum changes only $1.6\%$, from
$3.991$ for Masked FlowEdit to $4.053$ with both controls, indicating little
change in latent scale (Fig.~\ref{supp:fig:ce}(c,d)). All effects are mask-confined:
out-of-mask displacement is zero at every logged step.

\subsection{Visualization and Discussion}
\label{supp:sec:visualization-discussion}

\paragraph{Qualitative results.}
Fig.~\ref{supp:fig:additional-qualitative} provides six qualitative examples of
full-image scene-text translation into Chinese. Across different source languages
and text layouts, FlowTranslator replaces the selected source text with the
requested Chinese translation while preserving the surrounding scene and the local
appearance of the text region. These cases are consistent with the trajectory
diagnostics above: the controls strengthen the edit within the detected text mask,
while the background remains unchanged in latent space.

\FloatBarrier
\endgroup

\end{document}